\documentclass[letterpaper,journal]{IEEEtran}
\usepackage{amsmath,amsfonts,amssymb}
\usepackage{algorithmic}
\usepackage{algorithm}
\usepackage{array}
\usepackage[caption=false,font=normalsize,labelfont=sf,textfont=sf]{subfig}
\usepackage{textcomp}
\usepackage{stfloats}
\usepackage{url}
\usepackage{verbatim}
\usepackage{graphicx}
\usepackage{cite}
\usepackage{booktabs}
\usepackage{balance}
\usepackage{xcolor}
\usepackage[normalem]{ulem}
\newif\ifshowrevisions
\showrevisionstrue

\usepackage{tikz}
\usetikzlibrary{arrows.meta,calc,positioning}

\definecolor{stlcgblue}{HTML}{356A9A}
\definecolor{maskorange}{HTML}{D9822B}
\definecolor{scangreen}{HTML}{16836D}
\definecolor{figink}{HTML}{263238}
\definecolor{figsoft}{HTML}{F6F7F8}

\begin{document}

\title{ScanSTL: Parallel Robustness Evaluation for \\ Signal Temporal Logic}

\author{Gokhan Alcan
\thanks{Gokhan Alcan is with the Automation Technology and Mechanical Engineering, Tampere University, 33100 Tampere, Finland \linebreak E-mail: {\tt\small gokhan.alcan@tuni.fi}.}
\thanks{This work was supported by the Research Council of Finland Project under Grant 370881 and by the NVIDIA Academic Grant Program.}
}



\maketitle

\begin{abstract}
Repeated evaluation and differentiation of Signal Temporal Logic (STL) robustness can become a computational bottleneck in robot planning and control.
Sequential temporal recurrences limit parallelism, while dense masking increases memory requirements.
We propose ScanSTL, which combines associative temporal aggregation with parallel scans and ordered block reductions.
Eventually and Always use range extrema, while inclusive strong Until composes compact segment representations.
A common range engine handles bounded and shifted intervals, including the guards required before delayed Until witnesses.
Each exact temporal operator computes complete robustness traces with linear work and storage and logarithmic parallel depth on uniformly sampled finite signals.
An open source JAX implementation supports automatic differentiation, batching, and compilation.
We compare ScanSTL with STLCG and STLCG++ using CPU and GPU operator benchmarks and nine composed specifications.
Across these nine specifications at 512 samples, ScanSTL achieves geometric mean speedups of $243\times$ for forward evaluation and $104\times$ for gradient computation over STLCG++ in JAX on the CPU.
On an RTX~5090 GPU, ScanSTL evaluates unbounded Until over more than two million samples with median times below $0.1$\,ms for forward evaluation and $0.25$\,ms for gradient computation.
Simulated escort and patrol experiments with a robot dog further demonstrate faster repair of violating plans and greater solver capacity in model predictive control.
Supplementary materials, videos and codes can be found at project website: 
\texttt{https://alcanlab.github.io/ScanSTL/}
\end{abstract}

\section{Introduction}
\label{sec:introduction}

\IEEEPARstart{I}{n} many robotic applications, successful task execution requires satisfying conditions on both motion and the times at which events occur.
A robot may need to \textit{always} avoid obstacles, \textit{eventually} visit regions within deadlines and remain near a person \textit{until} the person reaches safety.
Signal Temporal Logic (STL) expresses such requirements over real-valued signals~\cite{maler2004}.
Its quantitative semantics assign a robustness value whose sign indicates satisfaction or violation and whose magnitude provides a margin~\cite{fainekos2009,donze2010}.
This makes a specification useful for verifying and optimizing trajectories.

STL has consequently been incorporated into model predictive control (MPC)~\cite{raman2014} and optimization using smooth robustness objectives~\cite{pant2017,gilpin2021}. 
Automatic differentiation propagates robustness through robot dynamics to control inputs~\cite{leung2023stlcg}.
Each optimizer iteration evaluates the specification and its gradient, while nested temporal operators also require sub-formula values at multiple starting times. 
Complete robustness traces are therefore central to this computation, whose cost grows with horizon and sampling resolution.

\begin{figure}[t!]
\centering
\begin{tikzpicture}[
  x=1mm, y=1mm,
  >={Stealth[length=1mm,width=.7mm]},
  panel/.style={
    draw=figink!18, fill=figsoft, rounded corners=1.2mm,
    line width=.35pt
  },
  tag/.style={
    rounded corners=.8mm, minimum width=14mm, minimum height=6mm,
    inner sep=1pt, font=\sffamily\bfseries\scriptsize, text=white
  },
  heading/.style={
    font=\sffamily\fontsize{6.2}{7.2}\selectfont, text=figink
  },
  rec/.style={
    circle, draw=stlcgblue, fill=white, minimum size=5mm,
    inner sep=0pt, font=\fontsize{6}{7}\selectfont
  },
  scanin/.style={
    rounded corners=.4mm, draw=scangreen!85, fill=white,
    minimum width=6mm, minimum height=2.8mm,
    inner sep=0pt, font=\fontsize{6}{7}\selectfont
  },
  scanop/.style={
    circle, draw=scangreen!85, fill=scangreen!12,
    minimum size=2.6mm, inner sep=0pt,
    font=\fontsize{6}{7}\selectfont
  },
  scanout/.style={
    rounded corners=.4mm, draw=scangreen, fill=scangreen!9,
    minimum width=6mm, minimum height=2.8mm,
    inner sep=0pt, font=\fontsize{6}{7}\selectfont
  },
  metric/.style={
    font=\sffamily\fontsize{6}{7.3}\selectfont,
    align=center, text=figink, text width=16mm, inner sep=0pt
  },
  flow/.style={
    ->, draw=stlcgblue!85, line width=.5pt,
    line cap=round, shorten <=.15mm, shorten >=.15mm
  },
  scanflow/.style={
    ->, draw=scangreen!85, line width=.5pt,
    line cap=round, shorten <=.15mm, shorten >=.15mm,
    preaction={draw=figsoft,line width=1.5pt,-}
  }
]
\path[use as bounding box] (0,0) rectangle (87,55);

\draw[panel] (.5,38.3) rectangle (86.5,54.5);
\node[tag,fill=stlcgblue] at (9,46.4) {STLCG};
\draw[figink!14] (18,39.8) -- (18,53);
\draw[figink!14] (68.5,39.8) -- (68.5,53);
\node[heading] at (43,52.2) {Sequential recurrence};
\foreach \i/\x in {0/25,1/37,2/49,3/61}{
  \node[rec] (h\i) at (\x,46.5) {$\rho_{\i}$};
  \node[font=\fontsize{6}{7}\selectfont,text=figink!85]
    (u\i) at (\x,40.7) {$p_{\i},q_{\i}$};
  \draw[flow] (u\i.north) -- (h\i.south);
}
\draw[flow] (h3.west) -- (h2.east);
\draw[flow] (h2.west) -- (h1.east);
\draw[flow] (h1.west) -- (h0.east);
\node[metric] at (77.5,46.4)
  {Serial depth\\[.7mm]$\mathcal O(T)$};

\draw[panel] (.5,20.6) rectangle (86.5,37.2);
\node[tag,fill=maskorange] at (9,28.9) {STLCG++};
\draw[figink!14] (18,22.1) -- (18,35.7);
\draw[figink!14] (68.5,22.1) -- (68.5,35.7);
\node[heading] at (43,35.2) {Parallel dense masks};
\fill[maskorange!30] (25,32.6) -- (33,32.6) -- (33,24.6) -- cycle;
\draw[step=1.6mm,draw=figink!40,line width=.25pt]
  (25,24.6) grid (33,32.6);
\draw[maskorange!90,line width=.45pt]
  (25,24.6) rectangle (33,32.6);
\node[heading,font=\sffamily\fontsize{5.8}{6.8}\selectfont]
  at (29,22.3) {Eventually / Always};
\foreach \dx/\dy in {2.4/2,1.2/1,0/0}{
  \path[fill=maskorange!10,draw=maskorange!85,line width=.3pt]
    ({48+\dx},{24.6+\dy}) rectangle ({55+\dx},{31.6+\dy});
  \draw[step=1.75mm,draw=figink!30,line width=.2pt]
    ({48+\dx},{24.6+\dy}) grid ({55+\dx},{31.6+\dy});
}
\fill[maskorange!30] (48,31.6) -- (55,31.6) -- (55,24.6) -- cycle;
\draw[step=1.75mm,draw=figink!40,line width=.25pt]
  (48,24.6) grid (55,31.6);
\draw[maskorange!90,line width=.45pt]
  (48,24.6) rectangle (55,31.6);
\node[heading,font=\sffamily\fontsize{5.8}{6.8}\selectfont]
  at (52.5,22.3) {Until};
\node[metric] at (77.5,28.9)
  {Storage\\[.6mm]$\mathcal O(T^2)$ unary\\[.6mm]$\mathcal O(T^3)$ Until};

\draw[panel] (.5,.5) rectangle (86.5,19.4);
\node[tag,fill=scangreen] at (9,9.95) {ScanSTL};
\draw[figink!14] (18,2) -- (18,17.9);
\draw[figink!14] (68.5,2) -- (68.5,17.9);
\node[heading] at (43,17.6) {Parallel suffix scan};
\foreach \i/\x in {0/25,1/37,2/49,3/61}{
  \node[scanin] (p\i) at (\x,14.2) {$P_{\i}$};
  \node[scanout] (r\i) at (\x,2.6) {$\rho_{\i}$};
}
\node[scanop] (q01) at (25,10.1) {$\otimes$};
\node[scanop] (q23) at (49,10.1) {$\otimes$};
\node[scanop] (s0) at (25,6.3) {$\otimes$};
\node[scanop] (s1) at (37,6.3) {$\otimes$};
\draw[scanflow] (p0.south) -- (q01.north);
\draw[scanflow] (p1.south west) -- (q01.north east);
\draw[scanflow] (p2.south) -- (q23.north);
\draw[scanflow] (p3.south west) -- (q23.north east);
\draw[scanflow] (q01.south) -- (s0.north);
\draw[scanflow] (q23.south west) -- (s0.north east);
\draw[scanflow] (p1.south) -- (s1.north);
\draw[scanflow] (q23.south west) -- (s1.north east);
\draw[scanflow] (s0.south) -- (r0.north);
\draw[scanflow] (s1.south) -- (r1.north);
\draw[scanflow] (q23.south) -- (r2.north);
\draw[scanflow] (p3.south) -- (r3.north);
\node[metric] at (77.5,9.95)
  {Work + storage\\$\mathcal O(T)$\\[.7mm]
   Parallel depth\\$\mathcal O(\log T)$};
\end{tikzpicture}
\caption{Comparison of temporal robustness evaluation.
STLCG propagates robustness through a sequential recurrence,
while STLCG++ evaluates temporal candidates using parallel dense masks.
ScanSTL combines compact segment representations through associative
scans, computing complete robustness traces with linear work and
storage and logarithmic parallel depth.}
\label{fig:overview}
\end{figure}
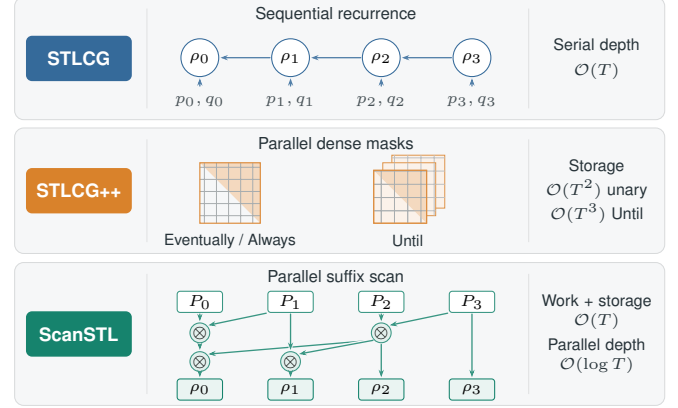

Existing differentiable evaluators expose a tradeoff between temporal dependencies and storage (Fig.~\ref{fig:overview}).
STLCG~\cite{leung2023stlcg} uses recurrences with sequential dependencies along the trace. 
STLCG++~\cite{kapoor2025stlcgpp} exposes parallelism through dense temporal masks. 
When the window grows with the trace, its storage is quadratic for Eventually and Always and cubic for strong Until.
Until is particularly demanding because each witness is combined with a condition that must hold up to that witness. 
These costs limit repeated evaluation within a planning or control budget.

To address this problem, we propose ScanSTL, a unified parallel evaluator for exact STL robustness.
ScanSTL combines associative temporal aggregation~\cite{mamouras2021} with parallel scans~\cite{blelloch1990} and ordered block reductions~\cite{vanherk1992,gil1993} to evaluate complete robustness traces without dense temporal masks or sequential recurrences along the full horizon.
Eventually and Always use range extrema, while inclusive strong Until composes compact segment representations.
A common range engine handles bounded and shifted intervals, including the conditions that must hold before an Until witness becomes eligible.
Each exact temporal operator requires linear work and storage and logarithmic parallel depth under explicit finite trace semantics.
For differentiable predicates, JAX supports automatic differentiation through the resulting piecewise differentiable computation without requiring smoothing.
Smooth Eventually and Always also retain this scan structure.

The main contributions of this work are:
\begin{enumerate}
\item A unified parallel evaluation algorithm for exact Eventually, Always, and inclusive strong Until, combining associative temporal aggregation with scans and range reductions. Each temporal operator computes complete robustness traces with linear work and storage and logarithmic parallel depth under finite-horizon semantics.
\item A common parallel range engine for bounded unary operators and Until pairs that preserves operand order and avoids duplicate contributions. It incorporates the guards required by shifted Until and also supports smooth unary reductions that are not idempotent.
\item An open-source JAX implementation of ScanSTL
supporting automatic differentiation, batching, and just-in-time compilation for use in optimization and control.
\item An evaluation across CPU and GPU operator benchmarks, nine composed STL specifications, and simulated robot planning and MPC tasks, quantifying computational gains in robustness evaluation, verified plan repair, and closed-loop solver capacity.
\end{enumerate}

\section{Related Works}
\label{sec:related}

\subsection{Control Using STL Robustness}

STL robustness guides control synthesis~\cite{fainekos2009,donze2010}.
Raman \textit{et al.}~\cite{raman2014} encoded STL constraints in mixed-integer MPC, while subsequent work reduced the number of binary variables~\cite{kurtz2022}. 
Direct robustness optimization supports nonlinear dynamics through smooth approximations~\cite{pant2017,gilpin2021}.
Arithmetic-geometric mean robustness accounts for satisfaction beyond the most critical sample~\cite{mehdipour2019}, robustness-to-go prevents past bottlenecks from obscuring differences between future plans~\cite{ilyes2026rotogo}, and GradSTL~\cite{chevallier2025} provides smooth STL semantics and derivatives for neurosymbolic reasoning and learning. 
These approaches develop control formulations or robustness objectives for optimization and learning.
ScanSTL addresses the complementary cost of repeatedly evaluating robustness and its gradients within optimization.

\subsection{Exact Monitoring and Associative Algorithms}

Early approaches to efficient STL monitoring reused computations across overlapping temporal windows.
Donz\'e \textit{et al.}~\cite{donze2013} combined temporal decompositions with sliding extrema algorithms~\cite{lemire2006}.
Deshmukh \textit{et al.}~\cite{deshmukh2017} developed robustness intervals for incomplete traces, and RTAMT~\cite{nickovic2020} translates bounded future formulas into past formulas for incremental monitoring in discrete and dense time. 
TLTk~\cite{cralley2020} parallelizes monitoring on CPUs and GPUs with linear trace storage for a fixed formula, but its reported time bound retains a temporal window factor. 
Rodionova \textit{et al.}~\cite{rodionova2016} interpreted Boolean MTL as filtering, with separate quantitative kernel scores that differ from spatial robustness.
Jak\v{s}i\'c \textit{et al.}~\cite{jaksic2018} compute semantic distances invariant to formula rewrites through weighted automata, whose size can be exponential in the formula. 
Mamouras \textit{et al.}~\cite{mamouras2021} established associative Since/Until pairs and ordered window aggregation with constant amortized work per sample. Their online monitor already achieves linear total work for a fixed formula, but does not develop input derivatives or logarithmic parallel depth across time.
ScanSTL adapts the local pairs to inclusive Until and combines Blelloch's scan and recurrence construction~\cite{blelloch1990} with block prefix and suffix reductions~\cite{vanherk1992,gil1993}. 

\subsection{Differentiable and Parallel STL Evaluation}

Leung \textit{et al.}~\cite{leung2023stlcg} introduced STLCG to express STL robustness as a computation graph, enabling backpropagation for trajectory optimization, parameter fitting, and learning with logical constraints.
Its temporal operators reuse intermediate results through recurrent updates, but the resulting dependencies limit parallel evaluation along the trace.
Kapoor \textit{et al.}~\cite{kapoor2025stlcgpp} developed STLCG++ to expose this temporal parallelism by applying masks to expanded signal tensors. 
The framework supports automatic differentiation in JAX and PyTorch.
This construction improves parallel utilization at the cost of larger intermediate arrays.
When temporal windows grow with the trace, storage is quadratic for Eventually and Always and cubic for Until, whose witnesses each require a prefix condition.
These two designs therefore highlight a tradeoff between sequential dependencies and memory use in differentiable robustness evaluation.
ScanSTL combines associative scans and ordered range reductions to compute each exact temporal operator with linear work and storage and logarithmic parallel depth, while retaining automatic differentiation in JAX~\cite{jax2018}.
We use STLCG and STLCG++ as baselines to evaluate how this formulation addresses the temporal dependencies of recurrence and the storage cost of dense masking.

\section{Problem Formulation}
\label{sec:problem}


Let $x = (x_0, \ldots, x_{T-1})$, $x_t \in \mathbb{R}^{d}$, be a uniformly sampled finite signal of length $T\geq 1$. 
An atomic predicate $\mu$ has a signed robustness function $r_\mu:\mathbb{R}^{d}\rightarrow\mathbb{R}$, for example $r_{h(x)\geq c}(x_t)=h(x_t)-c$.
Robustness is positive for satisfaction and negative for violation, while zero marks the boundary and its absolute value measures the margin.
We consider formulas
\begin{equation}
\varphi ::= \mu \mid \neg\varphi \mid \varphi_1\wedge\varphi_2
\mid \varphi_1\vee\varphi_2 \mid \Diamond_I\varphi \mid \Box_I\varphi
\mid \varphi_1\,\mathcal{U}_I\,\varphi_2,
\end{equation}
where $I=[a,b]$, $0\leq a\leq b$, has integer endpoints and permits $b=\infty$.
The operators $\Diamond$, $\Box$, and $\mathcal{U}$ denote Eventually, Always, and Until, respectively. 
Implication is derived as $\varphi_1\Rightarrow\varphi_2:=\neg\varphi_1\lor\varphi_2$. 
Weak Until and Release can likewise be derived and are not treated as primitives.


Let $\overline{\mathbb{R}}:=\mathbb{R}\cup\{-\infty,+\infty\}$. 
Evaluating a formula at every start time produces the robustness trace
\begin{equation}
\rho_\varphi\in\overline{\mathbb{R}}^T, \qquad
\rho_\varphi[t]:=\rho(\varphi,x,t),\quad t\in\{0,\ldots,T-1\}.
\label{eq:robustness-trace}
\end{equation}
The trace is written in natural time order, so index $t$ corresponds to evaluation beginning at signal index $t$.
Its time-zero entry $\rho_\varphi[0]$ is the scalar robustness of the complete signal. 
Every subformula similarly produces a length-$T$ trace consumed by its parent operator.

Since the signal is finite, temporal intervals are truncated at the right boundary.
At time $t$, define the effective upper endpoint and valid offset set by
\begin{align}
b_t^{\mathrm{eff}}&:=\min\{b,T-1-t\},
\label{eq:effective-upper-bound}\\
\mathcal{K}_t^{[a,b]}
&:=\left\{k\in\mathbb{Z}_{\geq0}\;\middle|\;
a\leq k\leq b_t^{\mathrm{eff}}\right\}.
\label{eq:valid-offsets}
\end{align}
Thus $b_t^{\mathrm{eff}}=T-1-t$ when $b=\infty$, and the valid set is empty when $a>b_t^{\mathrm{eff}}$. 
Empty reductions take the identity values $\max\emptyset=-\infty$ and $\min\emptyset=+\infty$.
Empty Eventually and Until intervals evaluate to $-\infty$, whereas an empty Always interval evaluates to $+\infty$. 
Truncation is performed entirely on indices and uses only available samples.


Writing $p_t:=\rho_{\varphi_1}[t]$ and $q_t:=\rho_{\varphi_2}[t]$, the exact finite-horizon quantitative semantics are
\begin{subequations}
\label{eq:exact-semantics}
\begin{align}
\rho_\mu[t] &= r_\mu(x_t),
\label{eq:sem-predicate}\\
\rho_{\neg\varphi_1}[t] &= -p_t,
\label{eq:sem-negation}\\
\rho_{\varphi_1\land\varphi_2}[t] &= \min(p_t,q_t),
\label{eq:sem-conjunction}\\
\rho_{\varphi_1\lor\varphi_2}[t] &= \max(p_t,q_t),
\label{eq:sem-disjunction}\\
\rho_{\varphi_1\Rightarrow\varphi_2}[t] &= \max(-p_t,q_t),
\label{eq:sem-implication}\\
\rho_{\Diamond_{[a,b]}\varphi_1}[t]
&=\max_{k\in\mathcal{K}_t^{[a,b]}}p_{t+k},
\label{eq:eventually-semantics}\\
\rho_{\Box_{[a,b]}\varphi_1}[t]
&=\min_{k\in\mathcal{K}_t^{[a,b]}}p_{t+k},
\label{eq:always-semantics}\\
\rho_{\varphi_1\,\mathcal{U}_{[a,b]}\,\varphi_2}[t]
&=\max_{k\in\mathcal{K}_t^{[a,b]}}
\min\left(q_{t+k},\min_{0\leq j\leq k}p_{t+j}\right).
\label{eq:until-semantics}
\end{align}
\end{subequations}
The Boolean operators in \eqref{eq:exact-semantics} are pointwise, while Eventually and Always are finite-horizon range maximum and minimum operations.
The Until definition is discrete-time \emph{strong} Until with the inclusive-overlap convention.
A valid witness $k\in[a,b_t^{\mathrm{eff}}]$ must satisfy $\varphi_2$ at $t+k$, while $\varphi_1$ must hold from $t$ through the witness time.
The inner prefix therefore includes both $j=k$ and, when $a>0$, all offsets $j<a$.
When no valid witness exists, the outer maximum is empty and the Until robustness is $-\infty$.


\textbf{Problem Statement:} Given the predicate robustness traces of an STL formula over a finite horizon $T$, the problem is to compute the complete robustness trace of the formula and each of its subformulas according to \eqref{eq:exact-semantics}.
The result must be obtained for every time index $t\in\{0,\ldots,T-1\}$, rather than only for the initial value $\rho_\varphi[0]$, while preserving the exact finite-horizon semantics, including interval truncation and empty-window identities.

A satisfactory solution should require low computational work and memory, expose parallelism across the temporal dimension with parallel depth substantially smaller than $T$, and avoid storage-intensive intermediate representations.
It should also remain compatible with automatic differentiation so that robustness gradients can be used in optimization and learning tasks.

\section{Proposed Method: ScanSTL}
\label{sec:method}

ScanSTL addresses the complete-trace problem in Sec.~\ref{sec:problem} by preserving the exact semantics in \eqref{eq:exact-semantics} and reorganizing their computation.
Each subformula produces a length-$T$ robustness trace, so a formula is evaluated recursively over its syntax tree, with every node returning a trace.
Boolean operators remain pointwise and require no further temporal construction.
The development therefore begins with unbounded Eventually and Always, where the associative-scan structure is visible directly.
Bounded ranges and strong Until are then built from the same associative viewpoint.

\subsection{Unbounded Eventually and Always}
\label{sec:unbounded-ea}

Let $p_t=\rho_{\varphi_1}[t]$. For $I=[0,\infty]$, the exact Eventually and Always traces are the suffix extrema
\begin{align}
\rho_{\Diamond_{[0,\infty]}\varphi_1}[t]
&=\max_{t\leq i<T}p_i=:s_t^{\max},\label{eq:suffix-max}\\
\rho_{\Box_{[0,\infty]}\varphi_1}[t]
&=\min_{t\leq i<T}p_i=:s_t^{\min}.\label{eq:suffix-min}
\end{align}
They satisfy the backward recurrences
\begin{align}
s_t^{\max}&=\max(p_t,s_{t+1}^{\max}),
&s_T^{\max}&=-\infty,\label{eq:suffix-max-recurrence}\\
s_t^{\min}&=\min(p_t,s_{t+1}^{\min}),
&s_T^{\min}&=+\infty.\label{eq:suffix-min-recurrence}
\end{align}
Because $\max$ and $\min$ are associative, each complete trace can be evaluated by one associative scan in reverse time rather than by following the corresponding recurrence sequentially.
An associative scan performs $\mathcal{O}(T)$ work with $\mathcal{O}(\log T)$ parallel depth.

For a shifted unbounded interval, the same suffix traces are read at $t+a$:
\begin{align}
\rho_{\Diamond_{[a,\infty]}\varphi_1}[t]
&=
\begin{cases}
s_{t+a}^{\max}, & t+a<T,\\
-\infty, & t+a\geq T,
\end{cases}
\label{eq:shifted-eventually}\\
\rho_{\Box_{[a,\infty]}\varphi_1}[t]
&=
\begin{cases}
s_{t+a}^{\min}, & t+a<T,\\
+\infty, & t+a\geq T.
\end{cases}
\label{eq:shifted-always}
\end{align}

\subsection{Bounded Range Reductions}
\label{sec:bounded-ranges}

For finite index bounds $\ell\leq r$, define the truncated range operators
\begin{align}
\operatorname{RangeMax}(p,\ell,r)
&:=\max_{i\in[\ell,r]\cap[0,T-1]}p_i,
\label{eq:range-max}\\
\operatorname{RangeMin}(p,\ell,r)
&:=\min_{i\in[\ell,r]\cap[0,T-1]}p_i.
\label{eq:range-min}
\end{align}
They return $-\infty$ and $+\infty$, respectively, when the truncated index set is empty.
The bounded unary semantics are therefore
\begin{align}
\rho_{\Diamond_{[a,b]}\varphi_1}[t]
&=\operatorname{RangeMax}(p,t+a,t+b),
\label{eq:eventually-range}\\
\rho_{\Box_{[a,b]}\varphi_1}[t]
&=\operatorname{RangeMin}(p,t+a,t+b).
\label{eq:always-range}
\end{align}

For a nonempty finite interval, set
\begin{equation}
\bar b:=\min\{b,T-1\}, \qquad w:=\bar b-a+1.
\label{eq:static-window}
\end{equation}
Here $\bar b$ is the largest offset that can be valid at $t=0$, whereas $b_t^{\mathrm{eff}}$ in \eqref{eq:effective-upper-bound} is the time-dependent truncation boundary.
If $a>\bar b$, every window is empty and its identity is returned directly.
Otherwise, identity padding turns \eqref{eq:eventually-range} and \eqref{eq:always-range} into fixed-width range queries of width $w$.

Consider any monoid $(\mathcal{M},\circ,e)$ and a sequence $z$ whose elements belong to $\mathcal{M}$.
Here, $\mathcal{M}$ is the set of values being combined, $\circ:\mathcal{M}\times\mathcal{M}\to\mathcal{M}$ is an associative combination rule, and $e\in\mathcal{M}$ is its identity, so $x\circ e=e\circ x=x$ for every $x\in\mathcal{M}$. 
After padding $z$ with $e$, partition it into blocks of width $w$.
Within every block, compute the left-to-right prefix values $F_i$ and the right-to-left suffix values $G_i$, while preserving the original operand order.
For a window beginning at $\ell=t+a$, let
$R_t:=z_\ell\circ\cdots\circ z_{\ell+w-1}$ denote its reduction.
The block summaries give
\begin{equation}
R_t=
\begin{cases}
G_\ell, & \ell\bmod w=0,\\
G_\ell\circ F_{\ell+w-1}, & \text{otherwise}.
\end{cases}
\label{eq:block-range}
\end{equation}
If the window begins at a block boundary, $G_\ell$ already represents the entire window and is used directly.
Otherwise, the window crosses exactly one block boundary.
Its reduction is obtained by composing $G_\ell$, the suffix of the first block, with $F_{\ell+w-1}$, the prefix of the second block.
These two summaries cover the window exactly once and preserve its temporal order.
Consequently, \eqref{eq:block-range} \textit{requires only associativity}.
It therefore applies to order-sensitive Until composition and to smooth range operators for which repeating a block would change the result.
This is why a single range engine can serve all these operators.

Padding beyond the signal with $e$ reproduces the finite-horizon truncation convention.
Choosing $(\circ,e)=(\max,-\infty)$ gives bounded Eventually, while $(\min,+\infty)$ gives bounded Always.
The within-block prefix and suffix scans operate in parallel across blocks, yielding $\mathcal{O}(T+\bar b)$ work and storage and $\mathcal{O}(\log w)$ parallel depth.

\subsection{Strong Until as Max--Min Pair Composition}
\label{sec:until-pair}

Strong Until is not a pointwise range maximum because each witness also depends on the survival prefix of $\varphi_1$. 

For the unbounded interval, let $h_t:=\rho_{\varphi_1\,\mathcal{U}_{[0,\infty]}\,\varphi_2}[t]$. Eq.~\eqref{eq:until-semantics} satisfies
\begin{equation}
h_t=\max\bigl(\min(p_t,q_t),\min(p_t,h_{t+1})\bigr),
\qquad h_T=-\infty.
\label{eq:until-recurrence}
\end{equation}
The recurrence can be viewed as the action of a \textit{max--min affine map}
\begin{equation}
f_{A,c}(z):=\max\bigl(c,\min(A,z)\bigr).
\label{eq:max-min-map}
\end{equation}
Associate each time index with the local pair
\begin{equation}
P_t=(A_t,c_t):=\bigl(p_t,\min(p_t,q_t)\bigr),
\label{eq:local-pair}
\end{equation}
so that \eqref{eq:until-recurrence} becomes $h_t=f_{A_t,c_t}(h_{t+1})$.

Distributivity of $\min$ over $\max$ on $\overline{\mathbb{R}}$ makes this family closed under composition. 
For two adjacent segments, with the earlier segment on the left, define
\begin{equation}
(A_1,c_1)\otimes(A_2,c_2):=
\left(
\min(A_1,A_2),
\max\bigl(c_1,\min(A_1,c_2)\bigr)
\right).
\label{eq:pair-composition}
\end{equation}
This pair represents the composed map $f_{A_1,c_1}\!\circ f_{A_2,c_2}$. 

Associativity follows directly by expanding three pairs.
For $P_i=(A_i,c_i)$, both parenthesizations give
\begin{equation}
\begin{aligned}
(P_1\otimes P_2)\otimes P_3
&=P_1\otimes(P_2\otimes P_3)\\
&=\Bigl(\min(A_1,A_2,A_3),\\[-0.2em]
&\quad\max\bigl(c_1,\min(A_1,c_2),
\min(A_1,A_2,c_3)\bigr)\Bigr).
\end{aligned}
\label{eq:pair-associativity}
\end{equation}
The identity is $e=(+\infty,-\infty)$. The operation is idempotent but not commutative, so every scan and range query must keep earlier times on the left.
The range engine in \eqref{eq:block-range} preserves this order and does not rely on idempotence.

The composed pair also has a direct semantic interpretation. For $\ell\leq r$, write
\begin{equation}
P_\ell\otimes\cdots\otimes P_r=(A_{\ell:r},C_{\ell:r}),
\end{equation}
where
\begin{align}
A_{\ell:r}
&=\min_{\ell\leq j\leq r}p_j,\label{eq:segment-a}\\
C_{\ell:r}
&=\max_{\ell\leq k\leq r}
\min\left(q_k,\min_{\ell\leq j\leq k}p_j\right).
\label{eq:segment-c}
\end{align}
Let $\pi_2(A,C):=C$ denote the projection onto the second coordinate of a pair.
The subscript $2$ indicates that the second entry is selected.
By \eqref{eq:segment-c}, this coordinate is exactly the strong-Until robustness whose survival prefix begins at $\ell$. In particular,
\begin{equation}
h_t=C_{t:T-1}
=\pi_2\bigl(P_t\otimes P_{t+1}\otimes\cdots\otimes P_{T-1}\bigr),
\label{eq:until-suffix-scan}
\end{equation}
so one associative suffix scan produces the complete unbounded Until trace with $\mathcal{O}(T)$ work and $\mathcal{O}(\log T)$ depth.

\subsection{Bounded and Shifted Until}
\label{sec:bounded-until}

For an interval whose first admissible witness is delayed by $a$ steps, the left formula must also hold before the pair composition begins.
Define the prefix guard
\begin{equation}
g_t^{(a)}:=
\begin{cases}
+\infty, & a=0,\\
\displaystyle\min_{0\leq j<a}p_{t+j}, & a>0,
\end{cases}
\label{eq:prefix-guard}
\end{equation}
using $+\infty$ padding beyond the signal. For a finite nonempty interval, let
\begin{equation}
C_t^{[a,b]}:=
\pi_2\bigl(P_{t+a}\otimes P_{t+a+1}\otimes\cdots
\otimes P_{t+\bar b}\bigr),
\label{eq:bounded-pair-range}
\end{equation}
Thus $C_t^{[a,b]}$ selects the strong-Until value from the composed pair over the admissible witness range.
Out-of-horizon pairs are filled with the pair identity.
Factoring the common guard through the outer maximum gives
\begin{equation}
\rho_{\varphi_1\,\mathcal{U}_{[a,b]}\,\varphi_2}[t]
=\min\bigl(g_t^{(a)},C_t^{[a,b]}\bigr).
\label{eq:bounded-until-method}
\end{equation}
The guard enforces $\varphi_1$ before the first allowed witness, while the second coordinate of the pair range enforces it from offset $a$ through each witness.
The finite pair range is evaluated by \eqref{eq:block-range} with $\circ=\otimes$.

For $[a,\infty]$, the same guard is combined with the shifted suffix result,
\begin{equation}
\rho_{\varphi_1\,\mathcal{U}_{[a,\infty]}\,\varphi_2}[t]
=\min\bigl(g_t^{(a)},h_{t+a}\bigr),
\label{eq:shifted-until-method}
\end{equation}
where $h_{t+a}=-\infty$ if $t+a\geq T$. The pair monoid is not invertible, so a bounded range cannot be recovered by cancelling two suffix summaries.
We use the ordered block construction in \eqref{eq:block-range} to answer these queries with linear work and logarithmic depth.
For $[a,\infty]$, one associative pair suffix scan combined with the prefix-guard range minimum produces the complete shifted Until trace with $\mathcal{O}(T)$ work and $\mathcal{O}(\log T)$ parallel depth.
For finite $[a,b]$, the associative block reduction and prefix guard produce the complete bounded Until trace with $\mathcal{O}(T+\bar b)$ work and $\mathcal{O}(\log(\bar b+1))$ parallel depth.
The depth includes both the witness range of width $w$ and the guard of width $a$, since $\max(a,w)\leq\bar b+1$.
Table~\ref{tab:operator-classification} summarizes the algebraic objects, evaluation strategies, and full-trace complexities for the exact Boolean and temporal operators.

\begin{table*}[t]
\caption{Exact ScanSTL Operator Classification and Full-Trace Complexity}
\label{tab:operator-classification}
\centering
\footnotesize
\setlength{\tabcolsep}{5pt}
\renewcommand{\arraystretch}{1.12}
\begin{tabular}{@{}llllll@{}}
\hline
Operator & Algebraic object & Evaluation & Work & Depth & Storage \\
\hline
$\neg,\land,\lor,\Rightarrow$ & pointwise lattice & elementwise
& $\mathcal{O}(T)$ & $\mathcal{O}(1)$ & $\mathcal{O}(T)$ \\
$\Diamond,\Box:\,[a,\infty]$ & max/min monoid & suffix scan
& $\mathcal{O}(T)$ & $\mathcal{O}(\log T)$ & $\mathcal{O}(T)$ \\
$\Diamond,\Box:\,[a,b]$ & max/min monoid & block range
& $\mathcal{O}(T{+}\bar b)$ & $\mathcal{O}(\log w)$
& $\mathcal{O}(T{+}\bar b)$ \\
$\mathcal{U}:\,[a,\infty]$ & pair monoid + guard & scan + range min
& $\mathcal{O}(T)$ & $\mathcal{O}(\log T)$ & $\mathcal{O}(T)$ \\
$\mathcal{U}:\,[a,b]$ & pair monoid + guard & block composition
& $\mathcal{O}(T{+}\bar b)$ & $\mathcal{O}(\log(\bar b{+}1))$
& $\mathcal{O}(T{+}\bar b)$ \\
\hline
\end{tabular}
\end{table*}

\subsection{Smooth Robustness}
\label{sec:smooth-robustness}

ScanSTL provides smooth trace reductions separately from the exact formula evaluator.
For $\beta>0$ and a nonempty set of finite values, define
\begin{align}
\operatorname{smax}_{\beta}(v_1,\ldots,v_n)
&:=\frac{1}{\beta}\log\sum_{i=1}^{n}\exp(\beta v_i),
\label{eq:smooth-max}\\
\operatorname{smin}_{\beta}(v_1,\ldots,v_n)
&:=-\operatorname{smax}_{\beta}(-v_1,\ldots,-v_n).
\label{eq:smooth-min}
\end{align}
The smooth maximum upper-bounds the exact maximum by at most $\log(n)/\beta$ and converges to it as $\beta\rightarrow\infty$.
The corresponding binary reduction is
\begin{equation}
u\oplus_\beta v
:=\frac{1}{\beta}\log\bigl(\exp(\beta u)+\exp(\beta v)\bigr),
\label{eq:smooth-binary}
\end{equation}
which is associative in real arithmetic and is evaluated with a stabilized log-add-exp primitive.
Smooth unbounded Eventually is therefore a suffix scan under $\oplus_\beta$, and bounded smooth Eventually uses the same block engine in \eqref{eq:block-range}.
The aligned-window branch is essential because $\oplus_\beta$ is not idempotent.
Smooth Always follows by sign duality.

The exact pair composition does not remain associative after replacing its minima and maxima by smooth surrogates, because \eqref{eq:pair-composition} relies on exact lattice distributivity.
Smooth Until is therefore provided as a reference surrogate.
It retains the exact candidate values
\begin{equation}
c_{k,t}:=\min\left(q_{t+k},\min_{0\leq j\leq k}p_{t+j}\right)
\label{eq:until-candidate}
\end{equation}
and smooths only the outer maximum over $k\in\mathcal{K}_t^{[a,b]}$.
This surrogate converges to exact Until as $\beta\rightarrow\infty$, but it is not equivalent to smoothing every nested minimum and maximum and is not presented as an associative fast path.
For finite $b$, the implementation constructs the survival prefixes for offsets $0,\ldots,\bar b$, giving $\mathcal{O}(T(\bar b+1))$ work and storage.
The unbounded reference path sets $\bar b=T-1$ and therefore has $\mathcal{O}(T^2)$ work and storage.
The convergence statements above concern reductions over valid samples.
The current smooth implementation uses finite padding to avoid undefined derivatives at padded entries.
Bounded unary scans use half the most negative representable value in the scaled log domain, while the Until reference uses $\pm10^9$.
These are numerical substitutes for identities only when padded contributions are negligible.
Extreme predicate values or temperatures can violate this assumption, so the smooth code does not provide unrestricted numerical guarantees.
The exact operators use the true infinite identities.

\begingroup
\setlength{\textfloatsep}{8pt plus 1pt minus 2pt}
\setlength{\dbltextfloatsep}{8pt plus 1pt minus 2pt}
\setlength{\floatsep}{6pt plus 1pt minus 2pt}
\setlength{\dblfloatsep}{6pt plus 1pt minus 2pt}
\section{Experiments}
\label{sec:experiments}

We address three questions:
\begin{itemize}
\setlength{\itemsep}{0pt}
\item \textbf{Q1:} How do robustness evaluation and differentiation scale
with trace length on CPUs and GPUs?
\item \textbf{Q2:} Do the gains persist across composed specifications?
\item \textbf{Q3:} Does faster evaluation reduce the time to a
verified satisfying plan and increase MPC solver capacity?
\end{itemize}

\subsection{Compared Methods and Protocol}
\label{sec:experiment-protocol}

ScanSTL, STLCG recurrence~\cite{leung2023stlcg}, and STLCG++ masking~\cite{kapoor2025stlcgpp} run in one JAX process, with \texttt{stljax}~\cite{stljax} baselines and unbounded Always using \texttt{padding=last}. 
Operator tests run on a Ryzen~9 9950X3D and RTX~5090.
We report synchronized medians after two warmups. 
Timing stops after 0.25 seconds and the minimum repeat count, or at the maximum count.
Operators use 3--50 calls and formulas 5--30.
Compilation is excluded unless stated otherwise.
Forward evaluation returns complete traces, and gradients differentiate $\rho[0]$ with respect to the inputs.
An independent oracle and tests check values, boundaries, compiled execution, and gradients.
Curve endpoints reflect runtime caps or analytical allocation guards, rather than measured memory failures.

\begin{figure*}[!t]
\centering
\includegraphics[width=0.9\textwidth]{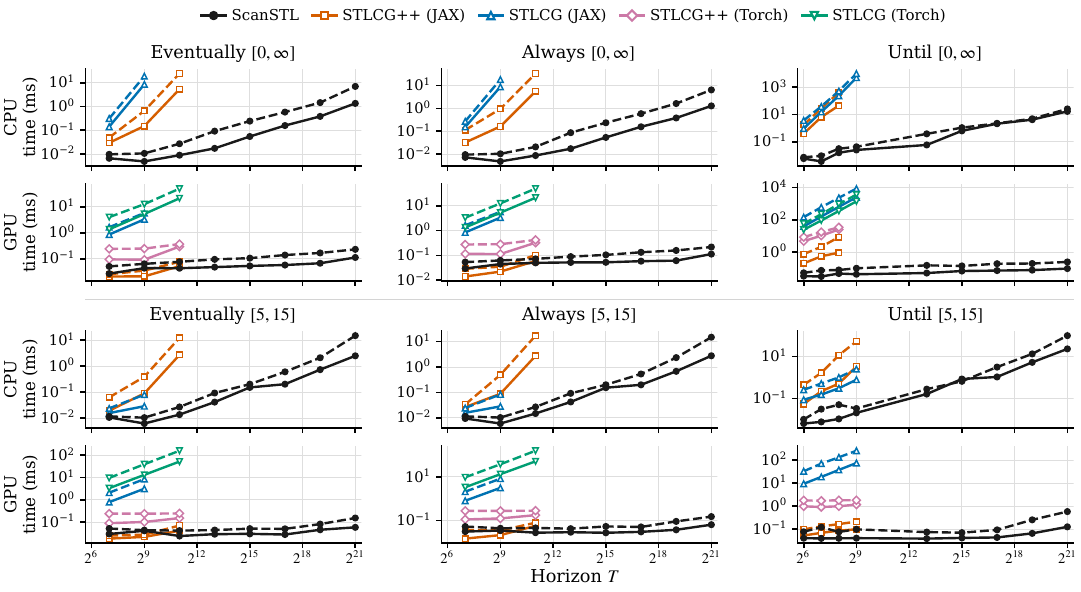}
\caption{Exact operator scaling for unbounded $[0,\infty]$ (rows 1--2) and bounded $[5,15]$ (rows 3--4) intervals, with Ryzen CPU above RTX~5090 GPU in each pair. Solid/dashed curves show forward traces/gradients of $\rho[0]$.}
\label{fig:exact-scaling}
\end{figure*}

\subsection{Operator Scaling (Q1)}
\label{sec:operator-results}

Figure~\ref{fig:exact-scaling} compares unbounded $[0,\infty]$ (rows 1--2) and bounded $[5,15]$ (rows 3--4) operators.

\textbf{Unbounded operators:} ScanSTL leads both baselines on the CPU at all shared lengths.
All reported speedups are relative to STLCG++ (JAX).

On the GPU, masking is faster for Eventually and Always at $T=128,512$.
At $T=2048$, ScanSTL gains $1.19$--$1.20\times$ forward, with gradients near parity for Eventually and $1.41\times$ faster for Always.

Since Until requires more involved reductions than unary operators to check survival through the witness, we compare it at the largest shared tested trace length.
For Until at $T=256$, ScanSTL gains $2869\times$ forward and $13{,}694\times$ for gradients on the CPU, with corresponding GPU gains of $19.7\times$ and $108\times$.

We have further evaluated these methods for the shifted interval $[5,\infty]$.
The results show similar speedup trends, as reported in Table~\ref{tab:shifted-speedups} with comparisons at the largest shared tested horizons, $T=2048$ for unary operators and $T=256$ for Until. At the longest tested horizon, $T=2^{21}$ on the GPU, ScanSTL evaluates unbounded Until in $0.0953$\,ms forward and $0.2442$\,ms for gradients.
Shifted Until takes $0.1303$ and $0.3854$\,ms, respectively.

\begin{table}[!t]
\caption{ScanSTL forward/gradient speedups over STLCG++ (JAX) for exact $[5,\infty]$ operators.}
\label{tab:shifted-speedups}
\centering
\footnotesize
\setlength{\tabcolsep}{4pt}
\begin{tabular}{@{}lrrr@{}}
\toprule
Operator & $T$ & Ryzen CPU ($\times$) & RTX 5090 ($\times$)\\
\midrule
Eventually & 2048 & 590/856 & 1.06/1.10\\
Always & 2048 & 619/1187 & 1.24/1.32\\
Until & 256 & 4035/7564 & 18.8/83.8\\
\bottomrule
\end{tabular}
\end{table}

\textbf{Bounded operators:} ScanSTL similarly leads on the CPU at all shared trace lengths.
Unary forward speedups at $T=2048$ are $185$--$201\times$ on the CPU and $1.73$--$2.06\times$ on the GPU.
Until at $T=512$ gains $160\times$ forward and $1528\times$ for gradients on the CPU, and $2.42\times$ and $2.17\times$ on the GPU.
We observed that the narrow window reduces masking cost, narrowing the GPU gap relative to unbounded Until.

\subsection{Composed Formulas (Q2)}
\label{sec:formula-results}

We evaluate the nine formulas in Table~\ref{tab:formula-suite} at $T\in\{32,64,$ $ 128, 256, 512\}$.
These formulas test nesting, bounded Until, response requirements, and multiple clauses.
The temporal intervals are $J_T=[0,T-1]$, $I_T=[0,\min(15,T-1)]$, and $W_T=[0,\min(128,T-1)]$.
We also define $r_i=p_{2i}\land p_{2i+1}$ and $\chi=\Diamond_{I_T}(r_2\land r_0)$.
For the remaining comparisons addressing Q2 and Q3, we use the JAX implementations of STLCG and STLCG++ because ScanSTL is also implemented in JAX.
The operator results in Fig.~\ref{fig:exact-scaling} further support this choice, as STLCG++ is faster in JAX at all shared test points, while STLCG performs similarly in JAX and PyTorch in most cases.

\begin{table}[!t]
\caption{Composite formula suite. Depth is maximum temporal nesting.}
\label{tab:formula-suite}
\centering
\footnotesize
\setlength{\tabcolsep}{3pt}
\renewcommand{\arraystretch}{1.10}
\begin{tabular}{@{}c c l l@{}}
\toprule
Spec. & Depth & Description & Formula\\
\midrule
$\varphi_1$ & 1 & Invariance & $\Box_{J_T}r_0$\\
$\varphi_2$ & 2 & Stabilization & $\Diamond_{J_T}\Box_{I_T}r_0$\\
$\varphi_3$ & 1 & Bounded Until & $p_0\,\mathcal{U}_{I_T}\,p_1$\\
$\varphi_4$ & 3 & Ordered visits & $\Diamond_{I_T}(r_4\land\Diamond_{I_T}(r_3\land\chi))$\\
$\varphi_5$ & 3 & Visit, then settle & $\Diamond_{I_T}(r_2\land\Diamond_{I_T}\Box_{I_T}r_0)$\\
$\varphi_6$ & 1 & Ten visits & $\bigwedge_{i=0}^{9}\Diamond_{I_T}r_i$\\
$\varphi_7$ & 2 & Response & $\Box_{[0,\infty]}(p_0\Rightarrow\Diamond_{W_T}p_1)$\\
$\varphi_8$ & 2 & Always Until & $\Box_{[0,\infty]}(p_0\,\mathcal{U}_{W_T}\,p_1)$\\
$\varphi_9$ & 1 & 32 visits & $\bigwedge_{i=0}^{31}\Diamond_{[0,\infty]}p_i$\\
\bottomrule
\end{tabular}
\end{table}

Figure~\ref{fig:formula-scaling} shows that ScanSTL is faster than both baselines for all nine formulas at each of the five tested horizons, in both forward evaluation and gradient computation.
Compared to STLCG++ (JAX), geometric mean speedups across the nine formulas increase from $6.21\times$ to $243\times$ for forward evaluation and from $5.68\times$ to $104\times$ for gradients as $T$ grows from 32 to 512.
The widening gap shows that the savings in temporal evaluation persist when operators are combined into complete specifications.
Compared to STLCG (JAX), geometric mean speedups similarly increase from $5.15\times$ to $140\times$ for forward evaluation and from $3.49\times$ to $57.5\times$ for gradients over the same horizons.

The size of the gain also depends on the formula structure.
For $\varphi_8$, which nests bounded Until inside Always, forward times at $T=512$ are $0.0267$, $182.6$, and $108.7$\,ms for ScanSTL, STLCG++, and STLCG, respectively. 
This composition particularly benefits from ScanSTL's efficient Until reductions.
At the same horizon, the more deeply nested unary formulas $\varphi_4$ and $\varphi_5$ achieve smaller forward speedups over STLCG of $4.42$--$4.64\times$.
Thus, nesting depth alone does not determine ScanSTL's advantage, as the temporal operators and their intervals also affect evaluation cost.

\begin{figure*}[!t]
\centering
\includegraphics[width=\textwidth]{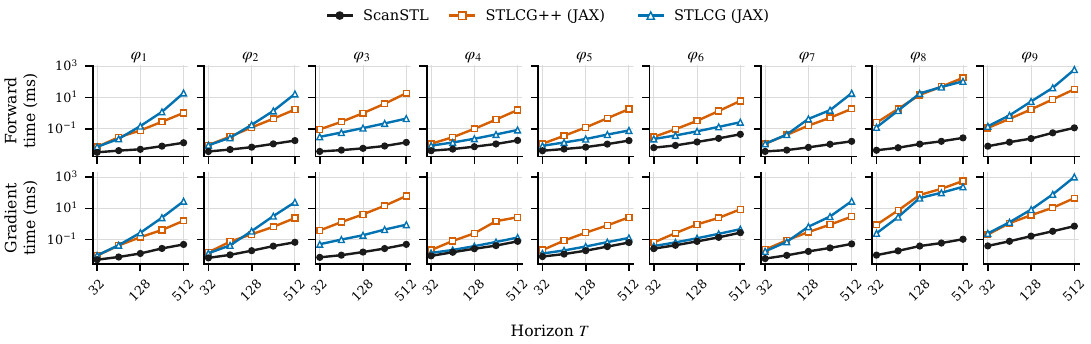}
\caption{Runtime scaling with trace length $T$ for the nine STL formulas in Table~\ref{tab:formula-suite}. Panels show median runtimes for complete robustness traces (top) and gradients of $\rho_\varphi[0]$ with respect to the input signals (bottom). All three methods use JAX on the i7-13700H CPU.}
\label{fig:formula-scaling}
\end{figure*}

\subsection{Escort-and-Patrol Planning and MPC (Q3)}
\label{sec:online-robot}

Finally, we test ScanSTL on a simulated robot dog with unicycle dynamics and $0.1$\,s sampling. 
The robot must remain within $1$\,m of a person until the person reaches a safe area, visit A and B within specified time windows, and return to its dock. 
Throughout the mission, it must stay inside a $6\times6$\,m arena and outside a \emph{robot exclusion region}, which the person may enter. 
We express these requirements as
\begin{equation}
\begin{aligned}
\varphi_{\mathrm{dog}}={}&
\mathrm{near}\;\mathcal{U}_{[0,\lfloor0.70T\rfloor]}\;\mathrm{humanSafe}\\
&\land\Diamond_{[\lfloor0.55T\rfloor,\lfloor0.80T\rfloor]}\mathrm{inA}\\
&\land\Diamond_{[\lfloor0.70T\rfloor,\lfloor0.92T\rfloor]}\mathrm{inB}\\
&\land\Box_{[0,\infty]}(\mathrm{robotAllowed}\land\mathrm{insideArena})\\
&\land\Diamond_{[\lfloor0.85T\rfloor,\infty]}\mathrm{inDock}.
\end{aligned}
\label{eq:robot-mission}
\end{equation}
The predicates use signed distance margins, with $\mathrm{robotAllowed}$ positive outside the exclusion region. 
Planning uses a known person trajectory, whereas MPC predicts it during execution.

\begin{table}[!t]
\caption{Time to an exactly satisfying plan with JAX after compilation (ms),
reported as mean $\pm$ sample standard deviation across three runs,
including optimization and verification.}
\label{tab:plan-repair-coverage}
\centering
\footnotesize
\setlength{\tabcolsep}{3pt}
\begin{tabular}{@{}lccc@{}}
\toprule
Method & $T=200$ & $T=300$ & $T=500$\\
\midrule
ScanSTL & $5.33\pm0.15$ & $5.78\pm1.92$ & $6.76\pm1.71$\\
STLCG & $529.38\pm16.98$ & $1942.58\pm812.12$ & $11{,}066.46\pm3631.67$\\
STLCG++ & $461.80\pm10.34$ & Compilation timeout & Not attempted\\
\bottomrule
\end{tabular}
\end{table}

\textbf{Time to a satisfying plan:} To isolate evaluation cost, all methods repair the same plans with matched exact temporal clauses, dynamics, and regularization in JAX.
Adam (learning rate $0.005$) maximizes a smooth minimum ($\beta=10$) of the five exact clause values.
Table~\ref{tab:plan-repair-coverage} reports the time to a satisfying plan after compilation, including optimization and verification.
ScanSTL's mean time increases from $5.33$ to $6.76$\,ms as $T$ grows from 200 to 500, while STLCG's increases from $529$\,ms to $11.07$\,s.
STLCG++ requires $462$\,ms at $T=200$ and compilation exceeds $120$\,s at $T=300$, and $T=500$ is therefore not attempted.
At $T=300$, the escort Until has 211 witness offsets.
STLCG++ explicitly constructs and reduces these masked cases, with $211\times511\times300=32{,}346{,}300$ entries per stacked tensor, or approximately $259$\,MB in float64.
ScanSTL performs prefix and suffix scans over three blocks of 211 pairs, followed by 300 pair compositions.
Each scan output contains only 633 pairs, or $10.1$\,kB in float64.

\textbf{Replanning during execution:} The MPC controller instead predicts $H$ person positions using a constant velocity Kalman filter.
Each tick performs 20 Adam updates from the previous solution and executes the first control.
The objective combines local smooth STL ($\beta$$=$$12$, weight $0.15$) with tracking, target, safety, smoothness, and effort terms, using the quadratic smooth Until (Sec.~\ref{sec:smooth-robustness}).
We evaluate 36 runs across $H\in\{30,60,120,240\}$, three seeds, and three methods, with the person following a zigzag trajectory that passes through the robot exclusion region.
Figure~\ref{fig:robot-compute} shows ScanSTL's median solver time rising from $1.40$ to $24.4$\,ms.
Within the nominal $100$\,ms solver budget, ScanSTL supports a prediction horizon of $24$\,s compared with $6$\,s for STLCG++ and $3$\,s for STLCG, based on the median solver times at the tested lookaheads.

\begin{figure}[!t]
\centering
\includegraphics[width=0.8\columnwidth]{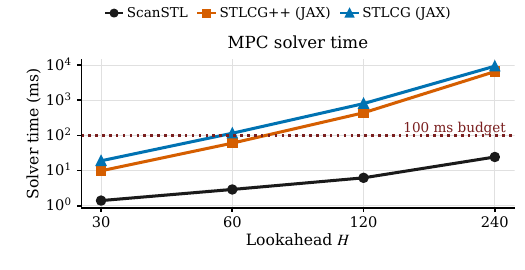}
\vspace{-10pt}
\caption{MPC solver times on i7-13700H with JAX. Curves and bands show
medians and interquartile ranges across three runs per method and lookahead.
The dotted line marks the nominal $100$\,ms control budget.}
\label{fig:robot-compute}
\end{figure}

Figure~\ref{fig:robot-run} illustrates how ScanSTL completes the mission with $H$$=$$30$.
The robot escorts the person while staying outside the robot exclusion region, then visits A and B and returns to the dock (panel~(a)).
Panel~(b) shows the local predicted STL cost, with shaded intervals marking the active task.
The peaks near task transitions reflect changes in the local planning objective, while mission satisfaction is assessed by exact verification of the executed trajectory.
Panel~(c) shows the minimum of the safety and proximity margins through the verified escort witness at $15.7$\,s, followed by the safety margin alone.
This margin remains positive throughout execution, with a minimum of $+0.091$\,m.
The remaining witnesses certify visits to A at $18.8$\,s, B at $21.4$\,s, and the dock at $25.6$\,s within their required intervals.
Together, these checks establish satisfaction of the complete mission with robustness $+0.091$.

\begin{figure*}[!t]
\centering
\includegraphics[width=0.95\textwidth]{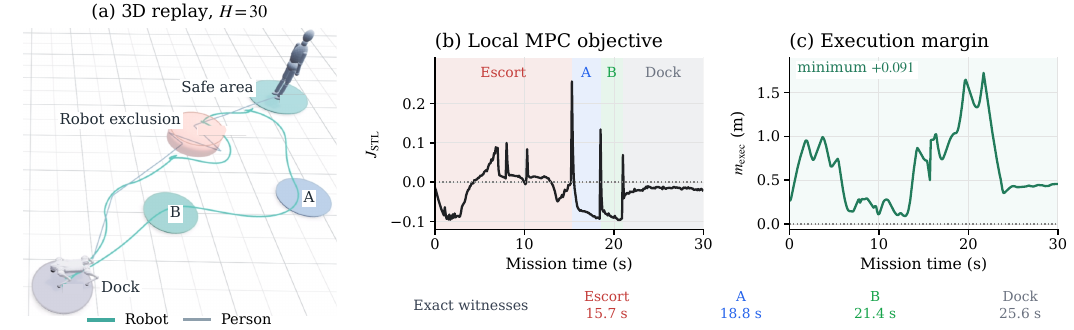}
\caption{A satisfying ScanSTL execution with $H=30$. (a) 3D replay.
(b) Local predicted STL cost, shaded by active task. (c) Exact safety and
escort margin. The maximizing temporal witnesses are listed below.}
\label{fig:robot-run}
\end{figure*}

\endgroup
\section{Conclusion}
\label{sec:conclusion}

We presented ScanSTL, a unified evaluator that combines associative scans and ordered range reductions for exact STL robustness.
The formulation computes complete traces for Eventually, Always, and inclusive strong Until with bounded and shifted intervals, requiring linear work and storage and logarithmic parallel depth per operator.
Its JAX implementation supports automatic differentiation for optimization and control.

The benchmarks demonstrate substantial computational gains, particularly for Until and composed specifications.
Simulated robot experiments show faster verified plan repair and greater MPC solver capacity.
Smooth Until remains a separate quadratic reference.
Future work will investigate efficient smooth Until evaluation and controllers that use the computational savings to improve task satisfaction under enforced deadlines.


\begin{thebibliography}{99}
\bibliographystyle{IEEEtran}

\bibitem{maler2004}
O. Maler and D. Ni\v{c}kovi\'c, ``Monitoring Temporal Properties of Continuous Signals,'' \textit{International Conferences on Formal Modeling and Analysis of Timed Systems and Formal Techniques in Real-Time and Fault-Tolerant Systems}, 152--166, 2004.

\bibitem{fainekos2009}
G. E. Fainekos and G. J. Pappas, ``Robustness of Temporal Logic Specifications for Continuous-Time Signals,'' \textit{Theoretical Computer Science}, 410(42), 4262--4291, 2009.

\bibitem{donze2010}
A. Donz\'e and O. Maler, ``Robust Satisfaction of Temporal Logic over Real-Valued Signals,'' \textit{International Conference on Formal Modeling and Analysis of Timed Systems}, 92--106, 2010.

\bibitem{raman2014}
V. Raman, A. Donz\'e, M. Maasoumy, R. M. Murray, A. Sangiovanni-Vincentelli and S. A. Seshia, ``Model Predictive Control with Signal Temporal Logic Specifications,'' \textit{IEEE Conference on Decision and Control}, 81--87, 2014.

\bibitem{pant2017}
Y. V. Pant, H. Abbas and R. Mangharam, ``Smooth Operator: Control Using the Smooth Robustness of Temporal Logic,'' \textit{IEEE Conference on Control Technology and Applications}, 1235--1240, 2017.

\bibitem{gilpin2021}
Y. Gilpin, V. Kurtz and H. Lin, ``A Smooth Robustness Measure of Signal Temporal Logic for Symbolic Control,'' \textit{IEEE Control Systems Letters}, 5(1), 241--246, 2021.

\bibitem{leung2023stlcg}
K. Leung, N. Ar\'echiga and M. Pavone, ``Backpropagation Through Signal Temporal Logic Specifications: Infusing Logical Structure into Gradient-Based Methods,'' \textit{The International Journal of Robotics Research}, 42(6), 356--370, 2023.

\bibitem{kapoor2025stlcgpp}
P. Kapoor, K. Mizuta, E. Kang and K. Leung, ``STLCG++: A Masking Approach for Differentiable Signal Temporal Logic Specification,'' \textit{IEEE Robotics and Automation Letters}, 10(9), 9240--9247, 2025.

\bibitem{mamouras2021}
K. Mamouras, A. Chattopadhyay and Z. Wang, ``Algebraic Quantitative Semantics for Efficient Online Temporal Monitoring,'' \textit{International Conference on Tools and Algorithms for the Construction and Analysis of Systems}, 330--348, 2021.

\bibitem{blelloch1990}
G. E. Blelloch, ``Prefix Sums and Their Applications,'' Technical Report CMU-CS-90-190, Carnegie Mellon University, 1990.

\bibitem{kurtz2022}
V. Kurtz and H. Lin, ``Mixed-Integer Programming for Signal Temporal Logic With Fewer Binary Variables,'' \textit{IEEE Control Systems Letters}, 6, 2635--2640, 2022.

\bibitem{mehdipour2019}
N. Mehdipour, C.-I. Vasile and C. Belta, ``Arithmetic-Geometric Mean Robustness for Control from Signal Temporal Logic Specifications,'' \textit{American Control Conference}, 1690--1695, 2019.

\bibitem{ilyes2026rotogo}
R. Ilyes, L. Bruderm\"uller, N. Hawes and B. Lacerda, ``Ro-To-Go! Robust Reactive Control with Signal Temporal Logic,'' \textit{IEEE International Conference on Robotics and Automation}, 7641--7648, 2026.

\bibitem{chevallier2025}
M. Chevallier, F. Smola, R. Schmoetten and J. D. Fleuriot, ``GradSTL: Comprehensive Signal Temporal Logic for Neurosymbolic Reasoning and Learning,'' \textit{International Symposium on Temporal Representation and Reasoning}, 6:1--6:14, 2025.

\bibitem{donze2013}
A. Donz\'e, T. Ferr\`ere and O. Maler, ``Efficient Robust Monitoring for STL,'' \textit{International Conference on Computer Aided Verification}, 264--279, 2013.

\bibitem{lemire2006}
D. Lemire, ``Streaming Maximum-Minimum Filter Using No More than Three Comparisons per Element,'' \textit{Nordic Journal of Computing}, 13(4), 328--339, 2006.

\bibitem{deshmukh2017}
J. V. Deshmukh, A. Donz\'e, S. Ghosh, X. Jin, G. Juniwal and S. A. Seshia, ``Robust Online Monitoring of Signal Temporal Logic,'' \textit{Formal Methods in System Design}, 51(1), 5--30, 2017.

\bibitem{nickovic2020}
D. Ni\v{c}kovi\'c and T. Yamaguchi, ``RTAMT: Online Robustness Monitors from STL,'' \textit{International Symposium on Automated Technology for Verification and Analysis}, 564--571, 2020.

\bibitem{cralley2020}
J. Cralley, O. Spantidi, B. Hoxha and G. Fainekos, ``TLTk: A Toolbox for Parallel Robustness Computation of Temporal Logic Specifications,'' \textit{International Conference on Runtime Verification}, 404--416, 2020.

\bibitem{rodionova2016}
A. Rodionova, E. Bartocci, D. Ni\v{c}kovi\'c and R. Grosu, ``Temporal Logic as Filtering,'' \textit{International Conference on Hybrid Systems: Computation and Control}, 11--20, 2016.

\bibitem{jaksic2018}
S. Jak\v{s}i\'c, E. Bartocci, R. Grosu and D. Ni\v{c}kovi\'c, ``An Algebraic Framework for Runtime Verification,'' \textit{IEEE Transactions on Computer-Aided Design of Integrated Circuits and Systems}, 37(11), 2233--2243, 2018.

\bibitem{vanherk1992}
M. van Herk, ``A Fast Algorithm for Local Minimum and Maximum Filters on Rectangular and Octagonal Kernels,'' \textit{Pattern Recognition Letters}, 13(7), 517--521, 1992.

\bibitem{gil1993}
J. Gil and M. Werman, ``Computing 2-D Min, Median, and Max Filters,'' \textit{IEEE Transactions on Pattern Analysis and Machine Intelligence}, 15(5), 504--507, 1993.

\bibitem{jax2018}
J. Bradbury \textit{et al.}, ``JAX: Composable Transformations of Python+NumPy Programs,'' 2018. [Online]. Available: \url{http://github.com/jax-ml/jax}

\bibitem{stljax}
UW-CTRL, ``stljax,'' GitHub repository. [Online]. Available:
\url{https://github.com/UW-CTRL/stljax}

\end{thebibliography}
\end{document}